\documentclass[conference]{IEEEtran}
\usepackage{cite}
\ifCLASSINFOpdf
  \usepackage[pdftex]{graphicx}
  \DeclareGraphicsExtensions{.pdf,.jpeg,.png}
\else
\fi
\usepackage{amsmath}
\usepackage{longtable}
\usepackage{algorithmic}
\usepackage{placeins}
\usepackage{amssymb}% http://ctan.org/pkg/amssymb
\usepackage{pifont}% http://ctan.org/pkg/pifont
\usepackage{booktabs}
\usepackage[colorlinks]{hyperref}

\usepackage[pscoord]{eso-pic}% The zero point of the coordinate systemis the lower left corner of the page (the default).

\newcommand{\placetextbox}[3]{% \placetextbox{<horizontal pos>}{<vertical pos>}{<stuff>}
  \setbox0=\hbox{#3}% Put <stuff> in a box
  \AddToShipoutPictureFG*{% Add <stuff> to current page foreground
    \put(\LenToUnit{#1\paperwidth},\LenToUnit{#2\paperheight}){\vtop{{\null}\makebox[0pt][c]{#3}}}%
  }%
}%

\usepackage{etoolbox}
\makeatletter
\patchcmd{\@makecaption}
  {\scshape}
  {}
  {}
  {}
\makeatletter
\patchcmd{\@makecaption}
  {\\}
  {.\ }
  {}
  {}
\makeatother

\begin{document}
\begin{titlepage}
\textbf{\color{white} Demonstration of Vector Flow Imaging using Convolutional Neural Networks}
\placetextbox{0.5}{0.26}{\parbox{\textwidth}{
    “© 2018 IEEE. Personal use of this material is permitted. Permission from IEEE must be obtained for all other uses, in any current or future media, including reprinting/republishing this material for advertising or promotional purposes, creating new collective works, for resale or redistribution to servers or lists, or reuse of any copyrighted component of this work in other works.”\\
    
    *IEEE policy provides that authors are free to follow public access mandates to post accepted articles in funding agency repositories. When posting in a funding agency repository, the IEEE embargo period is 24 months. However, IEEE recognizes that posting requirements and embargo periods vary by funder, and IEEE authors may comply with requirements to deposit their accepted manuscripts in funding agency repositories where the embargo is less than 24 months.
}}

\end{titlepage}

\title{Demonstration of Vector Flow Imaging using Convolutional Neural Networks}

\author{\IEEEauthorblockN{Thomas Robins,
Antonio Stanziola,
Kai Riemer,
Peter D. Weinberg,
Meng-Xing Tang\IEEEauthorrefmark{2}}

\IEEEauthorblockA{Department of Bioengineering, Imperial College London, United Kingdom}
\IEEEauthorblockA{\IEEEauthorrefmark{2}Email: mengxing.tang@imperial.ac.uk}}

\maketitle

\begin{abstract} 
%These methods are however susceptible to aliasing issues and require additional post processing that increases the scan time.%

% Downside: not optimal for rotation 

Synthetic Aperture Vector Flow Imaging (SA-VFI) can visualize complex cardiac and vascular blood flow patterns at high temporal resolution with a large field of view. Convolutional neural networks (CNNs) are commonly used in image and video recognition and classification. However, most recently presented CNNs also allow for making per-pixel predictions as needed in optical flow velocimetry. To our knowledge we demonstrate here for the first time a CNN architecture to produce 2D full flow field predictions from high frame rate SA ultrasound images using supervised learning. The CNN was initially trained using CFD-generated and augmented noiseless SA ultrasound data of a realistic vessel geometry. Subsequently, a mix of noisy simulated and real \textit{in vivo} acquisitions were added to increase the network's robustness. The resulting flow field of the CNN resembled the ground truth accurately with an endpoint-error percentage between 6.5\% to 14.5\%. Furthermore, when confronted with an unknown geometry of an arterial bifurcation, the CNN was able to predict an accurate flow field indicating its ability for generalization. Remarkably, the CNN also performed well for rotational flows, which usually requires advanced, computationally intensive VFI methods. We have demonstrated that convolutional neural networks can be used to estimate complex multidirectional flow.

\vspace{5mm}

\end{abstract}

\IEEEpeerreviewmaketitle
 
\section{Introduction} 
Abnormalities in blood flow and complex flow patterns have been linked to the development of cerebrovascular and coronary heart disease \cite{Lusis00}. Conventional Doppler ultrasound based methods are angle-dependent and can only measure the axial-component of blood flow in straight vessels, making them unsuitable for quantifying complex flow patterns \cite{Goddi17}. Vector Flow Imaging (VFI) is largely angle-independent and capable of visualizing highly resolved spatio-temporal flow patterns in all directions \cite{jensen14}. Several methods of VFI have been proposed such as Directional Beamforming \cite{jensen02}, Transverse Oscillation \cite{jensen98}, Multi-angle Doppler Vector Projectile Imaging \cite{billy14} and Echo Particle Image Velocimetry (Echo-PIV) \cite{leow15}. Echo-PIV is based on tracking the speckle pattern of two consecutive B-Mode images through cross-correlation to form a vector displacement field. From the pixel displacement and the time between two frames, a full 2D velocity field can be created \cite{leow15}. In Synthethic Aperture (SA) imaging these techniques are further enhanced through a wider field of view of the diverging waves, which allow imaging of regions with a limited acoustic window, such as in transthoracic cardiac imaging \cite{Hoyos16}. However, VFI comes at the cost of high beamforming loads, angle dependent errors and aliasing artifacts and can be time-consuming in post-processing \cite{jensen17}.

\begin{figure}[t]
\centering
\includegraphics[scale=0.85]{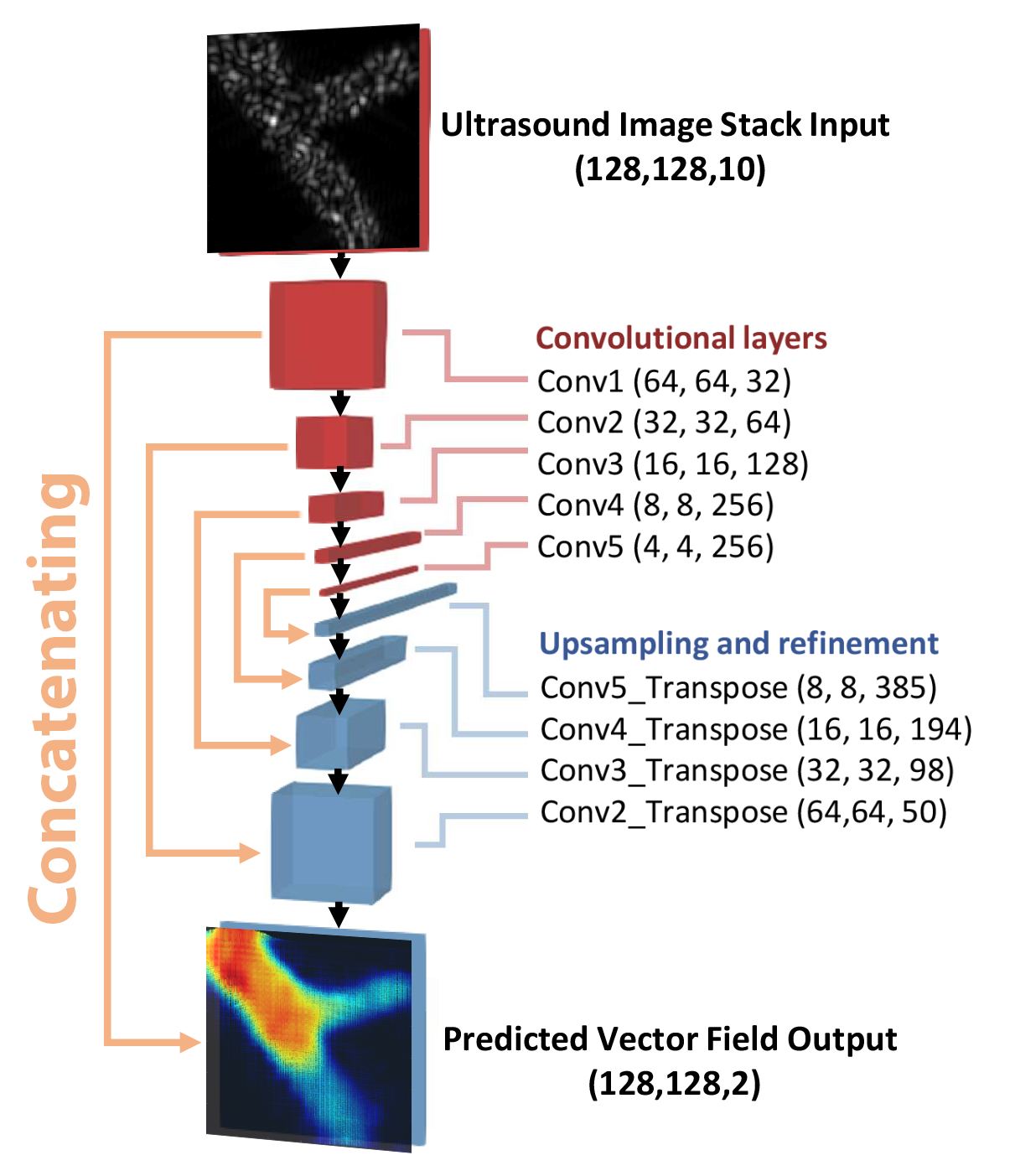}
\caption{Modified FlowNetSimple architecture for predicting 2D velocity fields from consecutive synthetic aperture ultrasound frames. }
\label{fig:Arch_diagram}
\end{figure} 

The problem of estimating 2D velocity fields from sets of consecutive images is widely explored in the computer vision community under Optical Flow estimation. Beside hand crafted models\cite{review_optical_flow}, machine learning approaches have also been used in this context, for example using principal component analysis of natural flow fields \cite{pcaflow} or neural networks \cite{deepflow,spatial_pyramid_flow}.
Among various neural network architectures, convolutional neural networks (CNNs) are a class for which each neuron of a hidden layer is connected to a local subset of neurons of the previous layer, and the weights of such connections are shared across the neurons of the same hidden layer. This allows the network to learn sets of hierarchical filters tailored for different characteristics of the data \cite{cnn_review}. CNNs have been used to estimate Optical Flow in a supervised manner using several variants of a convolutional architecture named FlowNet. Remarkably, FlowNet has developed as far as reaching real time optical flow detection up to 140 fps, outperforming state of the art methods \cite{flownet1, flownet2}.

In this study, we demonstrate the feasibility of using a CNN based on the FlowNetSimple architecture for velocity estimation of ultrasound images.
%\FloatBarrier\\

\section{Methods}
\subsection{Neural network}
Figure \ref{fig:Arch_diagram} illustrates the architecture of the modified FlowNetSimple CNN, where each convolutional layer but the last is using ReLU activation functions. The input of the network are image patches of 128 x 128 pixels and the output is a two channel image of the same size, where each channel represents a component of the 2D velocity vector. The size of the convolution filters is 5 pixels for the first layer and 4 pixels for all following layers, while between convolutional layers we perform a max-pooling operation over a neighbourhood of 2 by 2 pixels. This is followed by upsampling and refinement of the network with 4 by 4 kernels. Here we perform 'unpooling' and convolution while concatenating with feature maps from our convolutional layers, as seen in Fig.\ref{fig:Arch_diagram} (red to blue), to preserve high-level information through our CNN \cite{flownet1}.

\subsection{Training Data}
\begin{figure}[t]
\centering
\includegraphics[scale=0.4]{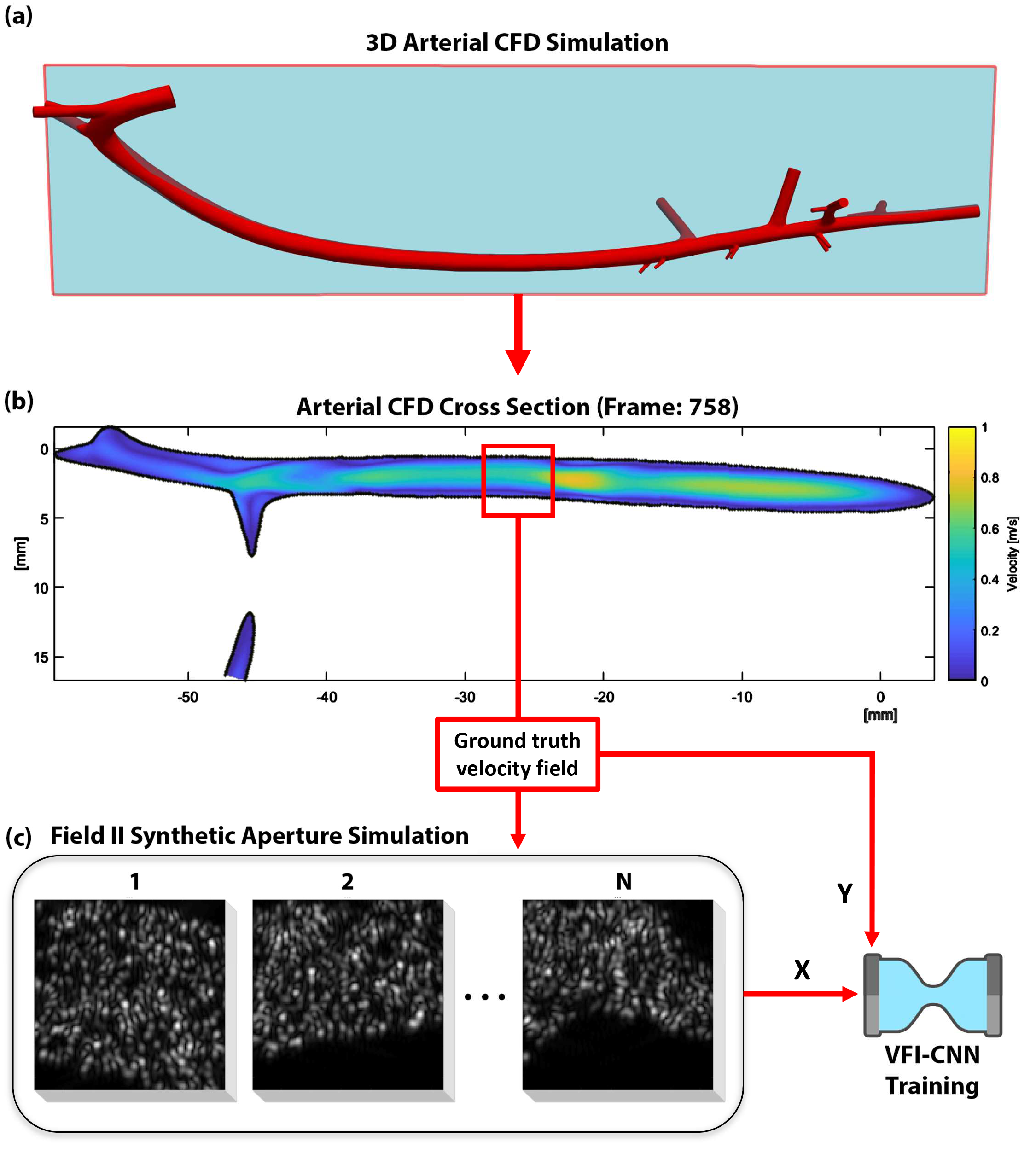}
\caption{Simulated synthetic aperture flow datasets from a 3D arterial CFD simulation. (a) £D high resolution rabbit artery model used and extracted cross section (marked in blue), (b) 2D velocity field sampled from the plane cross section of the 3D CFD simulation, (c) Simulated ultrasound datasets using randomly sampled CFD velocity field flow particles to be used for training the VFI-CNN}
\label{fig:Sampling}
\end{figure}

To learn how to perform displacement vector field estimation, the neural network requires a sufficiently large and diverse training dataset of vascular flow images. For each training dataset, a corresponding ground truth velocity field is required so that supervised learning can be performed. While ground truth information is difficult to determine for real world data, one solution is to generate simulated ultrasound images of vascular flow with known predetermined vector fields. This generated data, however, may not be sufficiently realistic to let the network generalize well for real world scenarios. Furthermore, it is computationally expensive and time consuming to build up large training sets of simulated data compared to sampling from external sources. An alternative approach is to use real world acquisitions and to attempt to approximate the ground truth using a gold standard VFI method. Consequently, to generate a versatile and large set of training data, both methods were used.\\

\subsubsection{Vascular Flow Simulation}
An artificial dataset of vascular flow was generated using StarCCM+ v11.02.01-R8 and a high-resolution luminal arterial surface of a 12-month-old male New Zealand White rabbit. Blood flow was prescribed time-dependent with rigid walls, Newtonian rheology and a no-slip boundary condition. To dissipate initial transients, three cycles over a total of 0.9 second divided in 1200 timesteps were simulated. Blood density was $\rho ρ = 1044.0 \frac{kg}{m^{3}}$ and dynamic viscosity was $\mu = 4.043\mathrm{e}{-3} \frac{Pa}{s}$. A physiological cardiac waveform and flow splits to the main branches were obtained from the literature \cite{Barakat97}. Velocity fields were saved for a number of 800x400px axial planar sections distributed along the vessel (Fig.\ref{fig:Sampling}a).\\

\subsubsection{Flow Field Window Sampling}
From these axial planar sections, velocity fields and flow particle positions were repeatedly extracted using a 128x128px region of interest with random position, orientation and point in time, for five consecutive time samples (Fig.\ref{fig:Sampling}b). \\

\subsubsection{Synthetic Aperture Image Generation}
Using the \textit{UltraSound ToolBox} \cite{utsb_web} and Field II (v3.24) \cite{jensen92,jensen_field:_1996} a time-resolved moving speckle phantom was created from the sampled flow field windows (Fig.\ref{fig:Sampling}c). To scan the phantom, a synthetic aperture setup with five virtual point sources was implemented. For the imaging sequence a 128-element linear array probe with a centre frequency of 8 MHz and 60\% bandwidth was modelled. In transmission a Hanning apodization was implemented. The excitation signal consisted of a tapered 3 cycle sinusoidal with a 50\% Tukey window and the pulse repetition frequency was 5 kHz. Using this approach, a training dataset of a total of 2400 unique B-Mode sequences was generated.\\     
\begin{table}[t]         
\centering
\caption{Data sets from IUS 2018 SA-VFI Challenge}
\begin{tabular}{ p{3.6cm}|p{1.5cm}p{1.5cm}   }
\toprule
Name                   & Simulation & Real \\
\midrule
\textbf{Carotid Bifurcation }   & \checkmark &                \\
Straight Vessel at 105 & \checkmark & \checkmark     \\
Straight Vessel at 90  & \checkmark & \checkmark     \\
Spinning Disk          & \checkmark &     \\
\bottomrule
\end{tabular}
\label{tabSAVFI}
\end{table}

\subsubsection{IUS 2018 SA-VFI Challenge Datasets}
To further increase the diversity of our training data and to make the CNN more robust, we also sampled the datasets provided by the IUS 2018 SA-VFI Challenge using the flow field window sampling (Table \ref{tabSAVFI}). As these came without velocity field information we generated a ground truth velocity data using our previously published gold standard Echo-PIV algorithm \cite{leow15}. Cross-correlation was performed using 32 by 32 pixel interrogation windows, which size was halved at each of the three iterations to refine the results. To demonstrate that our CNN is capable of generalizing, the Carotid Bifurcation model (Table \ref{tabSAVFI}) was excluded from the training data and was instead used exclusively for testing the performance of our CNN. An overview of the challenge training sets used can be seen in Table \ref{TrainData}.\\

\begin{table}[t]         
\centering
\caption{Overview of number of training, validation and testing datasets}
\begin{tabular}{ p{3.6cm}|p{1.0cm}p{1.0cm} p{1.0cm}   }
\toprule
Name                   & Training & Validation & Test\\
\midrule
Simulated CFD  & 1920 & 240  & 240 \\
iUS SA-VFi Challenge  & 4980 & 300  & 300\\
iUS SA-VFi Challenge SVD  & 5200 & 500  & 500\\
\bottomrule
\end{tabular}
\label{TrainData}
\end{table}

\subsubsection{In Line Data  Augmentation}
Due to the Hilbert transform applied during beamforming the training data is complex in nature. To use a real valued CNN, we extracted and concatenated the real and imaginary part of the input images, doubling the input data size. By inverting the complex intensity, we further augmented the training dataset, doubling its size again. Each image was normalized by its maximum absolute value. To extract moving blood speckle from tissue a Singular Value Decomposition filter (SVD) was applied. The datasets were evaluated with and without a binary vessel mask.  

\subsection{Error measure}
To evaluate our algorithm, we use a percentage endpoint error measure (EPE), which is defined as
\begin{equation}
    \textit{EPE} = 100\cdot \frac{E\left[\sqrt{(v_x - v'_x)^2+(v_y - v'_y)^2}\right]}{V_{max}}
\end{equation}
where $v_x$ and $v_y$ are the velocity components in axial and lateral dimension of the ground truth, $v'_x$ and $v'_y$ are the estimated components, $V_{max}$ is the maximum velocity magnitude in the image and $E[\cdot]$ is the sample average across all pixels.

\section{Results and Discussion}
\begin{figure}[t]
\centering
\includegraphics[width=0.49\textwidth]{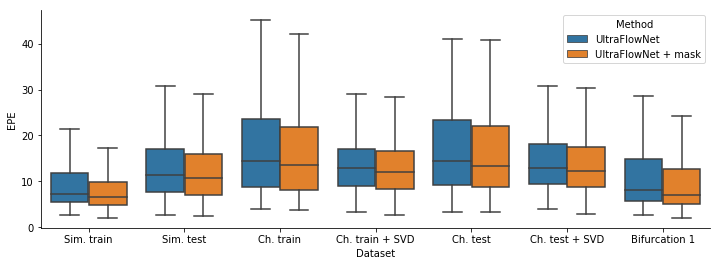}
\caption{Results obtained with different datasets. Note that the Bifurcation scan has not been used in training.}
\label{fig:results_boxplot}
\end{figure}

\begin{table}[]
    \centering
    \caption{Median EPE for different datasets}
    \begin{tabular}{ rcc } %p{3.6cm}|p{0.9cm} p{0.9cm}
    \multicolumn{3}{c}{EPE}\\
    \toprule
    Dataset & Train &Test\\
    \midrule
    Simulated CFD   &   7.3 & 11.5\\
    Simulated CFD + m  & 6.5 & 10.6\\
    \midrule
    Challenge   & 14.4    & 14.5 \\
    Challenge + m  &  13.6    & 13.2\\
    \midrule
    Challenge SVD   &  12.9    & 12.8\\
    Challenge SVD + m  & 12.1   & 12.2\\
    \midrule
    Bifurcation 1   & n.a.  & 8.1\\
    Bifurcation 1 + m  & n.a &  7.0\\
    \bottomrule
    \end{tabular}
    \label{tab:results}
\end{table}

While our method works well on simulated dataset, with a median EPE of 6.5\%, 
%We also observe that percentage endpoint-error doesnt increase significantly between training and testing datasets. 
we acknowledge that the highest error can be found for the unfiltered challenge datasets, with a median EPE of 14.5\%. We also observed, that the whiskers of the Tukey boxplots in Fig. \ref{fig:results_boxplot} extend several tens of percent above the median value, suggesting the presence of some flow fields for which the velocity field was not correctly estimated. \\

\begin{figure}[t]
\centering
\includegraphics[scale=0.26]{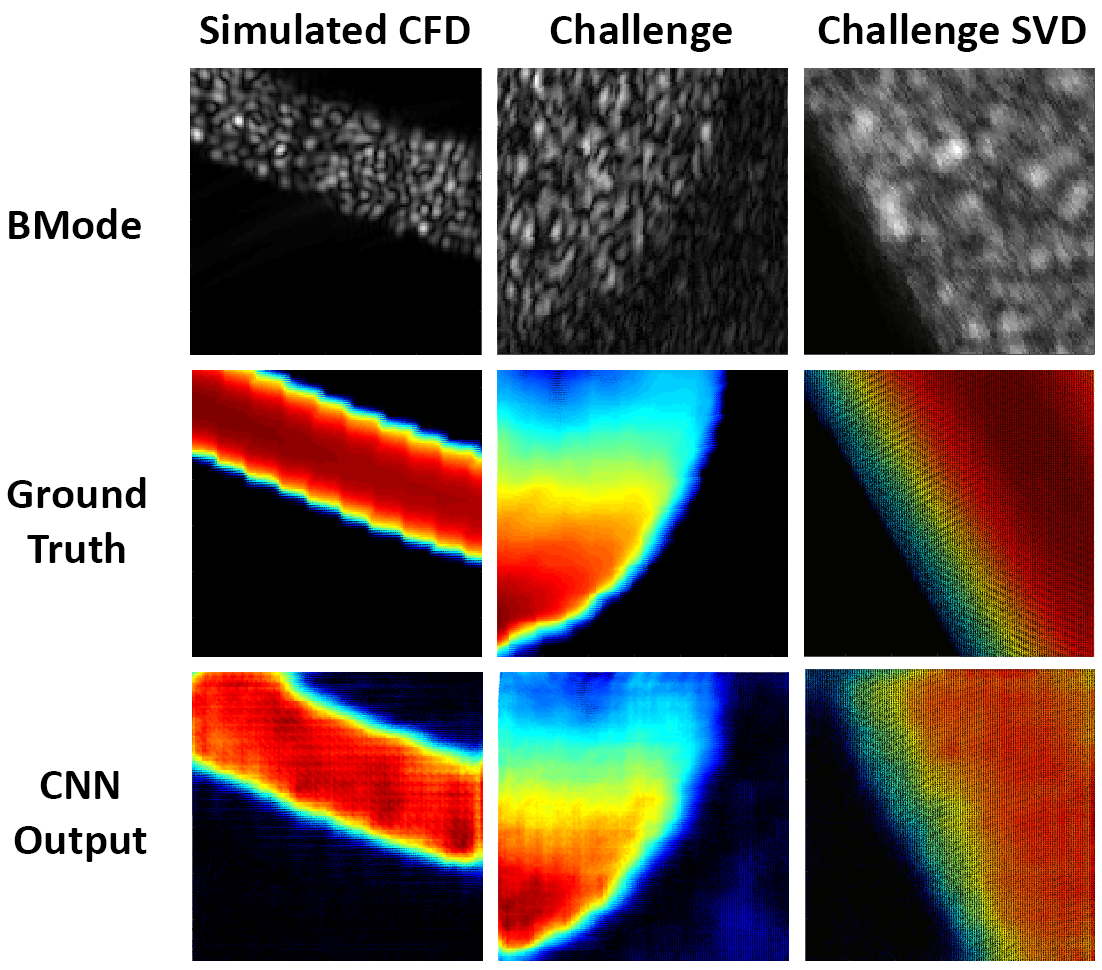}
\caption{Example of results using the proposed approach.}
\label{fig:Results}
\end{figure}

Areas of stagnation or zero movement were distinguished well from the flow field (Fig. \ref{fig:Results}). Rotational and straight flow is represented faithfully.\\ 

\begin{figure}[t]
\centering
\includegraphics[scale=0.24]{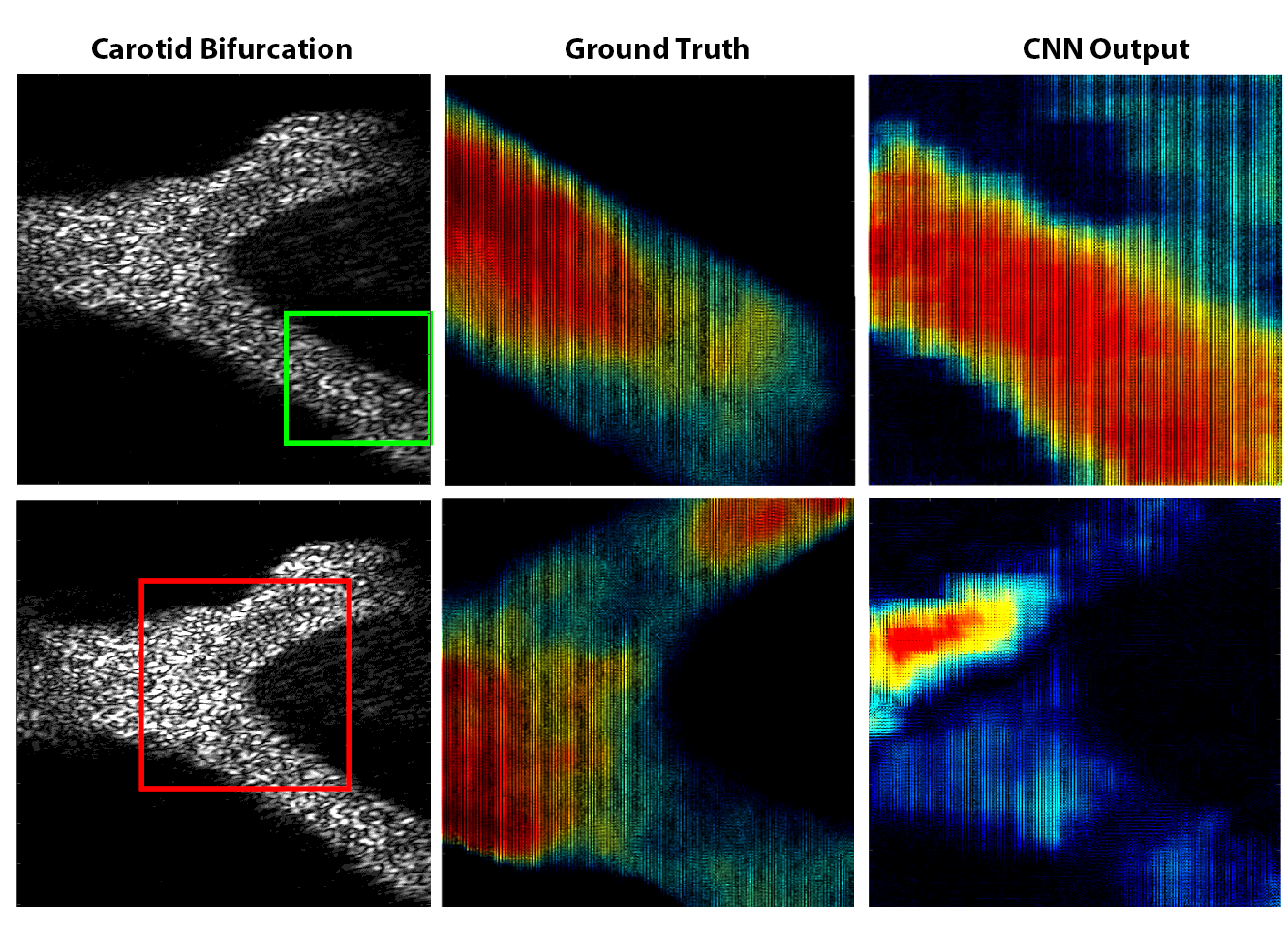}
\caption{Example of results for samples of the Carotid Bifurcation for regions of the External Carotid Artery (top) and Common Carotid Artery (bottom).}
\label{fig:BifurcationResults}
\end{figure}

When the network was applied to region 1 of the bifurcation model, which samples the External Carotid Artery (ECA), as shown in Fig. \ref{fig:BifurcationResults}, the result yielded a median EPE error of only 7.0\% (see Table \ref{tab:results}), despite the fact that the network had never been trained on a bifurcation phantom. However, when applied to the more complex flow of the Common Carotid Artery (CCA), the method did not accurately predict the velocity. \\

Future work should focus on the optimization of the neural network architecture, for example by adding a cross correlation stage in the architecture, as done for example in the more advanced FlowNet model \cite{flownet1,flownet2}. The training and evaluation must be performed on a wider range of diverse ultrasound data: the latter could be done for example by designing an unsupervised training approach, that could leverage different ultrasound scans without the need of a known ground truth velocity field. Also, regularization of the cost function, for example by imposing a smooth velocity field, should be evaluated to increase robustness. Lastly, the structure of the erroneously estimated flow fields should be carefully investigated to understand and circumvent the limitations of the approach.

\section{Conclusions}
In this study, we have shown that CNN can be used to estimate vector displacement fields from 2D flow ultrasound images. Furthermore, the preliminary results suggest that our approach is able to generalize the input data and to adapt to previously unknown geometries.

The use of CNNs could lead to real time evaluation of large 2D flow fields. In addition, it could be used to automatically extract clinically relevant information and to provide care teams with useful interpretations.  

\section{Acknowledgment}
We gratefully acknowledge the support of the NVIDIA Corporation for the their donation of a NVIDIA TITAN Xp GPU which was used to carry out this research. We would also like to acknowledge the BHF Centre of Research Excellence and the EPRSC for their funding. Finally, we thank the ULIS group for their support and Antonia Creswell for helpful feedback.

\end{document}